\documentclass[10pt,twocolumn,letterpaper]{article}

\usepackage{cvpr}
\usepackage{times}
\usepackage{epsfig}
\usepackage{graphicx}
\usepackage{amsmath}
\usepackage{amssymb}

\allowdisplaybreaks

\usepackage{booktabs}
\usepackage{array}
\usepackage{multirow}
\usepackage{tabularx}
\newcolumntype{Y}{>{\centering\arraybackslash}X}

\usepackage{graphbox}

\usepackage{url}

\DeclareMathOperator{\smpl}{SMPL}

\usepackage{xcolor}

\usepackage[pagebackref=true,breaklinks=true,letterpaper=true,colorlinks,bookmarks=false]{hyperref}

\cvprfinalcopy %

\def\cvprPaperID{1884} %

\ifcvprfinal\fi
\begin{document}

\title{Exploiting temporal context for 3D human pose estimation in the wild}

\author{Anurag Arnab\textsuperscript{1}\thanks{Equal contribution.} \hspace{0.01mm} \thanks{Work done during an internship at DeepMind}\\
	{\tt\small aarnab@robots.ox.ac.uk}
	\and
	Carl Doersch\textsuperscript{2}\footnotemark[1]\\
	{\tt\small doersch@google.com}
	\and
	Andrew Zisserman\textsuperscript{1,2}\\
	{\tt\small zisserman@google.com}
	\and 
	\textsuperscript{1}University of Oxford
	\quad
	\textsuperscript{2}DeepMind
}

\maketitle

\begin{abstract}
We present a bundle-adjustment-based algorithm for recovering accurate 3D human pose and meshes from monocular videos. 
Unlike previous algorithms which operate on single frames, we show that reconstructing a person over an entire sequence gives extra constraints that can resolve ambiguities.
This is because videos often give multiple views of a person, yet the overall body shape does not change and 3D positions vary slowly.
Our method improves not only on standard mocap-based datasets like Human 3.6M -- where we show quantitative improvements -- but also on challenging in-the-wild datasets such as Kinetics.
Building upon our algorithm, we present a new dataset of more than 3 million frames of YouTube videos from Kinetics with automatically generated 3D poses and meshes.
We show that retraining a single-frame 3D pose estimator on this data improves accuracy on both real-world and mocap data by evaluating on the 3DPW and HumanEVA datasets.
\end{abstract}

\section{Introduction}

Understanding the 3D configuration of the human body has numerous real-life applications in robotics, augmented and virtual reality, and animation, among other fields.
However, it is an inherently under-constrained problem when only a single image is available, as there are many 3D poses which project to the same 2D image.
Data-driven methods to resolve this ambiguity are promising, but they are typically trained and evaluated on motion capture datasets recorded in constrained and unrealistic environments \cite{ionescu_pami_2014, sigal_ijcv_2010, mehta_3dv_2017, joo_iccv_2015}.

To resolve some of the ambiguities in monocular 3D pose estimation, we exploit temporal consistency across frames of a video.
The temporal dimension of ordinary videos encodes valuable information:
multiple views of people are observed, where the body shape and bone lengths remain constant throughout a video, and joint positions in both 2D and 3D change slowly over time.
These priors constrain the space of possible poses and thus help reduce the ambiguity of this ill-posed problem as shown in Fig.~\ref{fig:teaser}.
Despite its value, the temporal information in mocap datasets is discarded by all current leading 3D pose estimation algorithms \cite{kanazawa_cvpr_2018, pavlakos_cvpr_2018, sun_eccv_2018, martinez_iccv_2017} which use only single, ambiguous frames.
Our approach incorporates temporal information through a form of {\em bundle adjustment}, a method used in multi-view geometry for estimating cameras and 3D structure of rigid 
scenes from image correspondences~\cite{Hartley04a,Triggs00b}.
We repurpose bundle adjustment to deal with non-rigid (articulated) human motion in a video sequence.
In contrast to previous recurrent models for human pose \cite{hossain_eccv_2018}, our method can jointly reason about all frames in the video, and errors made in initial frames do not accumulate over time.
As illustrated in Fig.~\ref{fig:teaser}, the current state-of-art single frame estimation network for the SMPL model \cite{kanazawa_cvpr_2018} fails on a number of frames of  ``in the wild'' videos, such as when there is occlusion, unusual poses, poor lighting or motion blur.
Our bundle adjustment method is able to correct these estimates and infer 3D human pose for these frames.

To address the lack of real-world data in 3D pose estimation, we apply our bundle adjustment framework to ``in the wild'' clips from the Kinetics dataset~\cite{kay_arxiv_2017} comprised of YouTube videos, and show how we can leverage our predictions on real-world videos as a source of weak supervision to improve existing 3D pose estimation models.
By encouraging temporal consistency with bundle adjustment and using YouTube videos as a source of weakly supervised data, we make the following novel contributions:

\begin{figure*}[t]

\def \imwidth {0.17\linewidth}
\def \imageid {369fccd180fe16bd_151_161}

\vspace{-\baselineskip}
\hspace*{-0.5cm}
\begin{tabular}{ m{1cm} ccccc}
Input                                                       &   
	\includegraphics[align=c, width=\imwidth]{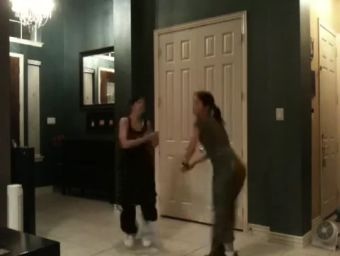} & 
	\includegraphics[align=c, width=\imwidth]{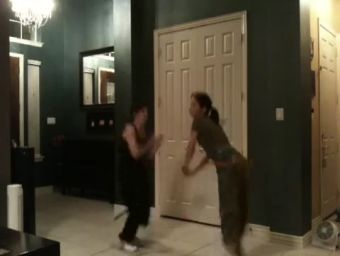} & 
	\includegraphics[align=c, width=\imwidth]{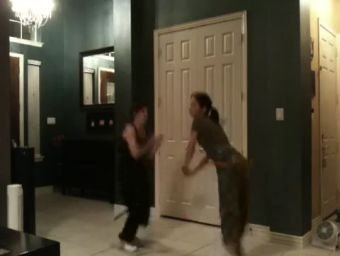} & 
	\includegraphics[align=c, width=\imwidth]{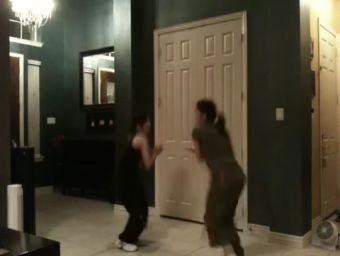} &
	\includegraphics[align=c, width=\imwidth]{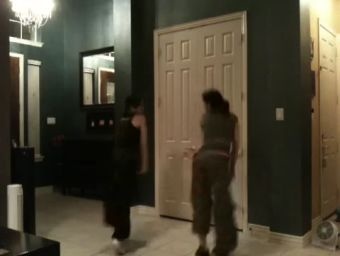} \\ \addlinespace[0.1cm]
Per-frame \cite{kanazawa_cvpr_2018}  &
	\includegraphics[align=c, width=\imwidth]{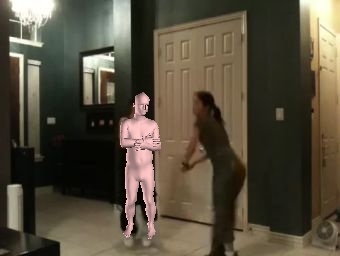} & 
	\includegraphics[align=c, width=\imwidth]{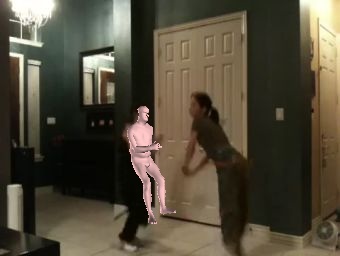} & 
	\includegraphics[align=c, width=\imwidth]{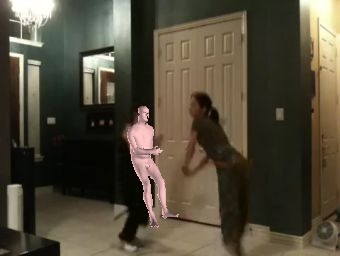} & 
	\includegraphics[align=c, width=\imwidth]{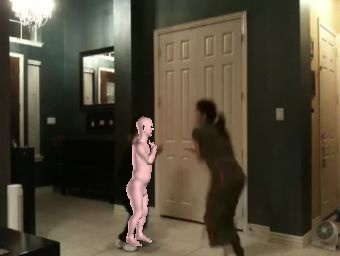} &
	\includegraphics[align=c, width=\imwidth]{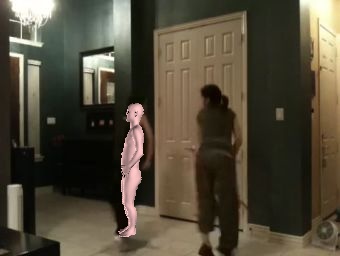} \\ \addlinespace[0.1cm]
Bundle adjustment &
	\includegraphics[align=c, width=\imwidth]{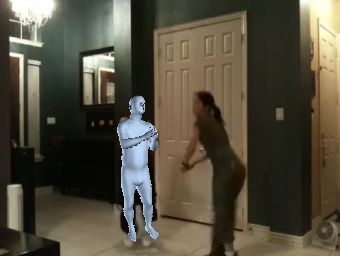} &
	\includegraphics[align=c, width=\imwidth]{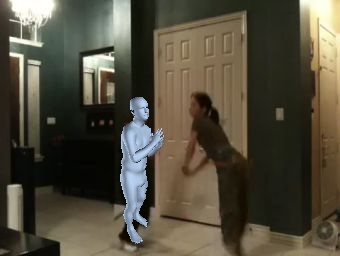} & 
	\includegraphics[align=c, width=\imwidth]{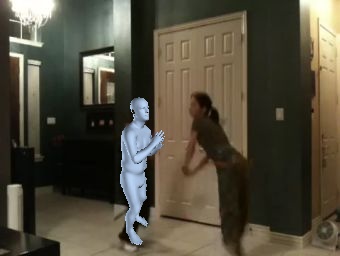} & 
	\includegraphics[align=c, width=\imwidth]{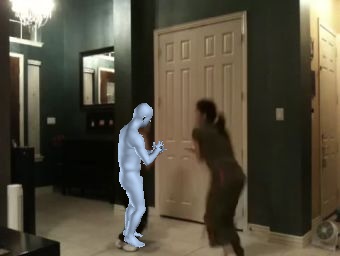} &
	\includegraphics[align=c, width=\imwidth]{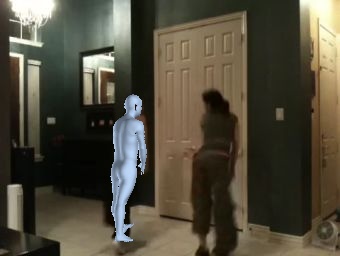} \\
& \multicolumn{5}{l}{
$\large \mathbf{\xrightarrow{\hspace*{16.3cm}}}$
}
\end{tabular}
\caption{
Although monocular 3D pose estimation is an ill-posed problem, state-of-art methods \cite{kanazawa_cvpr_2018} do not use temporal information to constrain the problem.
Coupled with the fact that 3D supervision is only available from lab-captured mocap datasets, they often fail on ``in the wild'' videos, e.g., from Kinetics \cite{kay_arxiv_2017}.
As shown in the second row, the failure modes of \cite{kanazawa_cvpr_2018} vary even though the image has barely changed.
Our proposed bundle adjustment considers all frames in the video jointly and uses temporal coherence to prevent major failures (column 2 and 3) and to resolve ambiguities (column 5).
We then apply our method on YouTube videos to obtain weakly-supervised data to improve per-frame methods. %
Note that we are only showing 5 out of 190 frames in the clip.
Best viewed in colour on screen.
}
\label{fig:teaser}
\vspace{-0.5\baselineskip}
\end{figure*}

First, we show that multi-frame bundle adjustment can be specialized to
human pose estimation, which improves performance on the Human 3.6M dataset over single frame estimation. 
Our method achieves the state-of-the-art for SMPL \cite{loper_tog_2015} models on this dataset.

We then apply our bundle adjustment method to 107 000 YouTube videos from the Kinetics dataset \cite{kay_arxiv_2017} and generate a large-scale dataset of 3D human poses aligned with the video frames.
This dataset contains great diversity in pose, with 400 different human actions, 
and is available publicly\footnote{\url{https://github.com/deepmind/Temporal-3D-Pose-Kinetics}}.
As we are fitting SMPL body models \cite{loper_tog_2015} to the data, other information such as 2D keypoints and body-part segmentations can also be obtained automatically as done by \cite{lassner_cvpr_2017}.

By retraining the single-frame 3D pose estimator using our automatically-generated dataset, we obtain a more robust network that performs better on real-world (3DPW \cite{von_marcard_eccv_2018}) and mocap (HumanEVA \cite{sigal_nips_2008}) datasets.
We are thus the first paper, to our knowledge, to show how we can use masses of unlabelled real-world data to improve 3D pose estimation models.

\section{Related Work}
3D human pose is typically represented in the literature as either a point cloud of 3D joint positions or the parameters of a body model.
A common approach with the former representation is to ``lift'' 2D keypoints (either ground truth or from a 2D pose detector) to 3D.
This has been recently done with neural networks \cite{martinez_iccv_2017, zhao_pami_2017, moreno_cvpr_2017} and previously using a dictionary of 3D skeletons \cite{ramakrishna_eccv_2012, akhter_cvpr_2015, zhou_cvpr_2016, wang_cvpr_2014} or other priors \cite{taylor_cviu_2000, valmadre_eccv_2010, akhter_cvpr_2015} to constrain the problem.
The point cloud representation also allows one to train a CNN to regress directly from an image (instead of 2D keypoints) to 3D joints using supervision from motion capture datasets like Human 3.6M \cite{pavlakos_cvpr_2017, sarandi_arxiv_2018, pavlakos_cvpr_2018b}.
However, this approach overfits to the constrained environments of lab-captured motion capture datasets and does not generalise well to real-world images.
Whilst methods based on ``lifting'' are more robust to this domain shift, they discard valuable information from the image as they depend solely on the input 2D keypoints.

Training models with supervision from both 2D keypoints (from real-world datasets such as \cite{lin_eccv_2014, andriluka_cvpr_2014, johnson_bmvc_2010}) and 3D joints (from mocap datasets) has been shown to help with generalisation to real-world images \cite{zhou_iccv_2017, rogez_cvpr_2017, mehta_3dv_2017, dabral_eccv_2018, sun_iccv_2017, sun_eccv_2018}.
However, greater success has been achieved in this scenario by fitting parametric models of human body meshes to images.
Human body models, such as \cite{loper_tog_2015} and \cite{anguelov_tog_2005}, encapsulate more prior knowledge, thus reducing the ambiguity of the 3D pose estimation problem.
Explicit priors such as bone length ratios remaining constant \cite{zhou_iccv_2017, dabral_eccv_2018} and limbs being symmetric \cite{dabral_eccv_2018} are enforced naturally by body models.
Moreover, this mesh representation also enables a direct mapping to body part segmentations \cite{lassner_cvpr_2017, pavlakos_cvpr_2018, kanazawa_cvpr_2018}.

Early work used the SCAPE body model \cite{anguelov_tog_2005} and fitted it to images using manually annotated keypoints and silhouettes \cite{guan_iccv_2009, sigal_nips_2008, balan_cvpr_2007, hasler_cvpr_2010}.
More recent works use the SMPL model \cite{loper_tog_2015} and fit it automatically.
This is done by either solving an optimisation problem to fit the model to the data \cite{bogo_eccv_2016, lassner_cvpr_2017, zanfir_cvpr_2018, balan_cvpr_2007} or by regressing the model parameters directly using a neural network \cite{kanazawa_cvpr_2018, omran_3dv_2018, pavlakos_cvpr_2018, tung_nips_2017} or random forest \cite{lassner_cvpr_2017}.
Optimisation-based approaches minimise an energy function that depends on the reprojection error of the 3D joints onto 2D \cite{bogo_eccv_2016, lassner_cvpr_2017}, priors on joint angle and shape parameters \cite{bogo_eccv_2016, lassner_cvpr_2017}, and/or the discrepancy between the silhouette of the 3D model and its foreground mask in the 2D image \cite{lassner_cvpr_2017, balan_cvpr_2007}.
Direct regression methods, in contrast, train a neural network where the keypoint \cite{kanazawa_cvpr_2018, omran_3dv_2018, pavlakos_cvpr_2018} or silhouette reprojection errors are used in its training objective \cite{pavlakos_cvpr_2018, omran_3dv_2018}.
Kanazawa~\etal~\cite{kanazawa_cvpr_2018} also use an adversarial loss that distinguishes between real and fake joint angles of SMPL models.
This effectively acts as a joint-angle prior, allowing the authors to utilise existing ground truth SMPL model fits from \cite{loper_tog_2014} without requiring them to be paired to images.

Our approach uses the per-frame neural network 
model of Kanazawa \etal \cite{kanazawa_cvpr_2018} as the initialisation of our optimisation problem.
Despite efforts by \cite{kanazawa_cvpr_2018} to train it with realistic 2D data, we show (as illustrated by Fig.~\ref{fig:teaser},~\ref{fig:diagram}) how this model often fails on challenging real-world videos and how these errors can be corrected with bundle adjustment.
Moreover, we show how we can improve the performance of this network by finetuning it using the results of our bundle adjustment as ground truth on originally difficult sequences.

We note that despite there being previous efforts to produce temporally consistent fits of the SMPL model \cite{huang_3dv_2017, zhang_uist_2018, zanfir_cvpr_2018, peng_tog_2018}, none of these works have been able to use these results to improve a per-frame model as we have.
Furthermore, \cite{zhang_uist_2018} and \cite{peng_tog_2018} have not explicitly evaluated on 3D pose estimation either.
Additionally, we do not assume knowledge of calibrated cameras like \cite{huang_3dv_2017, zanfir_cvpr_2018}.

There are also several methods which enforce temporal consistency without body models:
The works of \cite{gotardo_cvpr_2011, wandt_pami_2016, li_cvpr_2018} were based on Non-Rigid Structure from Motion whilst
\cite{andriluka_cvpr_2010} lifted tracked 2D keypoints into 3D.
More recently, Hossain \etal \cite{hossain_eccv_2018} also lifted 2D keypoints using an LSTM in a sequence-to-sequence \cite{sutskever_nips_2014} model. However, it is difficult to retain memory over long sequences as evidenced by their model performing best with a temporal context of only five frames.
Dabral \etal \cite{dabral_eccv_2018} use a feedforward network using the predictions of the previous 20 frames as input.
Our optimisation based approach, in contrast, can consider all frames (our experiments have as many as 1175 frames) in the video to produce more globally coherent results.
Furthermore, as we consider all frames jointly, rather than sequentially like \cite{dabral_eccv_2018, hossain_eccv_2018}, errors do not accumulate over time.

Finally, we note that there are several works which synthesise additional training data using rendering engines \cite{rogez_ijcv_2018, varol_cvpr_2017, chen_3dv_2016}.
Although this approach provides additional diversity compared to motion capture datasets, the resultant data, although fully labelled, is not photorealistic.
Our approach is complementary in that we leverage unlabelled, but %
real-world YouTube videos from the Kinetics dataset.
Concurrently to this paper, \cite{kanazawa_cvpr_2019} have also used additional videos from Instagram to improve 3D pose estimation models.
\section{Bundle Adjustment using the SMPL Model}
\label{sec:method_single_person}
\begin{figure*}
    \vspace{-\baselineskip}
	\centering
	\includegraphics[width=0.98\linewidth]{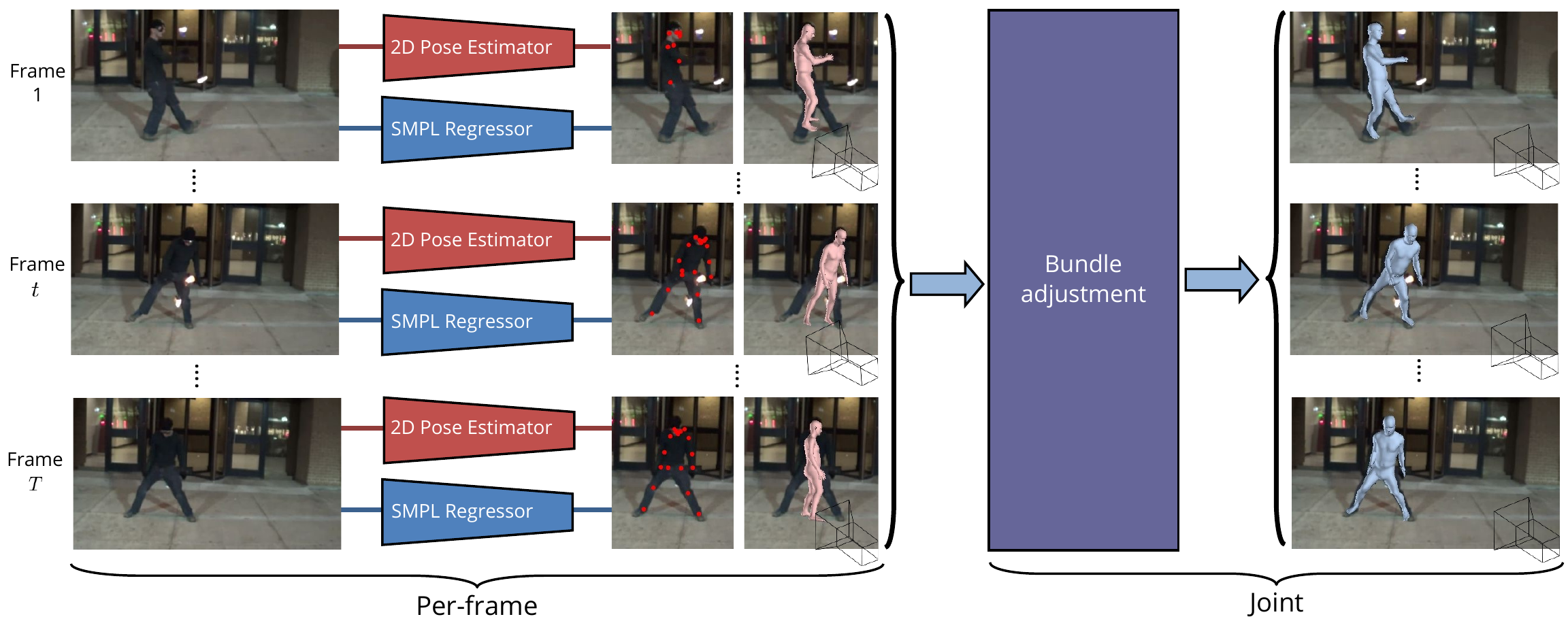}
	\caption{
	Overview of our method:
	Using initial per-frame estimates of 2D keypoints, SMPL- and camera parameters, we jointly optimise over the whole video comprising $T$ frames by encouraging temporal consistency.
	As a result, we can overcome poor 2D keypoint detection  (first row) and poor initial SMPL estimates (all rows) to output accurate SMPL- and camera-parameters.
	}
	\label{fig:diagram}
\end{figure*}

We jointly optimise the parameters of a SMPL statistical body shape model \cite{loper_tog_2015} and a camera over an entire video sequence. 
The whole-video approach contrasts with recurrent networks such as \cite{hossain_eccv_2018} which are only effective using a temporal context of only five frames, and allows for better global solutions.
As shown in Fig.~\ref{fig:diagram}, the input to our method is a sequence of video frames, 2D keypoint predictions for a single person for each frame using a state-of-art 2D pose detector~\cite{papandreou_cvpr_2017} and initial SMPL parameters produced per-frame using the HMR network of \cite{kanazawa_cvpr_2018}. 
From this, our method outputs SMPL- and camera parameters for each frame in the video that are consistent with each other and reproject to the 2D keypoints. 
In Sec.~\ref{sec:body_model}, we briefly describe the SMPL body model that we are fitting to videos. 
Thereafter, in Sec.~\ref{sec:formulation}, we detail the objective function that we minimise in order to fit this model to the video. 
Section.~\ref{sec:optimisation} we provide details on the optimisation.

\subsection{Body representation}
\label{sec:body_model}

The SMPL body model \cite{loper_tog_2015} parameterises a triangulated mesh with $N = 6890$ vertices of a human body.
It factorises the 3D mesh into shape parameters, $\mathbf{\beta} \in \mathbb{R}^{10}$ and pose $\mathbf{\theta} \in \mathbb{R}^{3K}$, where $K = 23$ joints.
The shape parameters model the variations in body proportions, height and weight.
They are the coefficients of a low-dimensional shape space that was originally learned by \cite{loper_tog_2015, bogo_eccv_2016} from a training set of approximately 4000 registered human-body scans.
The pose parameters model the deformation of the body as a result of the articulation of its $K$ internal joints.
They are an axis-angle representation of the relative rotation of a joint with respect to its parent in the model's kinematic tree.
SMPL is a differentiable function that outputs a mesh and positions of joints in 3D.
We denote the latter as $\mathbf{X} = \smpl(\beta, \theta) \in \mathbb{R}^{J \times 3}$ where $J$ is the number of joints.

\subsection{Formulation}
\label{sec:formulation}

We optimise an objective function that considers the reprojection of 3D keypoints onto 2D, temporal consistency of SMPL parameters, 3D- and 2D-keypoints, and a prior:
\begin{equation}
    E(\mathbf{\beta}, \mathbf{\theta}, \mathbf{\Omega}) = E_{R}(\mathbf{\beta}, \mathbf{\theta}, \mathbf{\Omega}) + E_{T}(\mathbf{\beta}, \mathbf{\theta}, \mathbf{\Omega}) + E_{P}(\theta, \beta)
    \label{eq:energy}
\end{equation}

\paragraph{Reprojection error:}
We assume that we have 2D keypoint detections, $\mathbf{x}_{det, i}$ with a confidence score of $w_i$ for the $i^{th}$ joint. 
This error term penalises deviations of the projections of our estimated 3D joints onto 2D over all $T$ frames in the video for all $J$ body joints:
\begin{align}
    & E_R(\mathbf{\beta}, \mathbf{\theta}, \mathbf{\Omega}) = \lambda_R\sum_{t}^{T}\sum_{i}^{J}{ w_i \rho(\mathbf{x}^{t}_{i} - \mathbf{x}^{t}_{det, i}) }.
    \label{eq:frame_energy}
\end{align}
Here, $\rho$ is the robust Huber error function which we favour over a squared error as it can deal better with noisy estimates that we sometimes obtain on ``in-the-wild'' sequences, and the superscript $t$ denotes time.
$\mathbf{x}$ is the 2D projection of the 3D joint $\mathbf{X}$,
\begin{align}
    \mathbf{x}^{t} & = s^t \Pi(R\mathbf{X}^{t}) + u^t \\
    \mathbf{X}^{t} & = \smpl(\mathbf{\beta}, \mathbf{\theta}^{t}), 
    \label{eq:smpl}
\end{align}
where $\Pi$ is an orthographic projection, $R \in \mathbb{R}^{3 \times 3}$ is the global rotation matrix parameterised by a Rodrigues vector and $\mathbf{\Omega}^t = \{s^t, u^t\}$ are the camera parameters comprising of scale, $s \in \mathbb{R}$ and translation $u \in \mathbb{R}^{2}$ and time-step $t$. 
Note that the parameters $\mathbf{\beta}$ and $\mathbf{\theta}$ are mapped to 3D joint positions $\mathbf{X}$ by $\smpl$, and that we use a single $\mathbf{\beta}$ parameter for the whole sequence as the body shape of the video's subject remains constant.

\paragraph{Temporal error:}
This error, $E_T$ is defined as:
\begin{align}
    E_{T}(\mathbf{\beta}, \mathbf{\theta}, \mathbf{\Omega}) = \sum_{t = 2}^{T} \sum_{i=1}^{J} & \lambda_1 \rho(\mathbf{X}^t_i - \mathbf{X}^{t-1}_i) + \lambda_2 \rho(\mathbf{x}^t_i - \mathbf{x}^{t-1}_i) \nonumber \\
    & + \lambda_3\rho(\mathbf{\Omega}^t - \mathbf{\Omega}^{t-1}).
    \label{eq:temporal}
\end{align}
The temporal error on 3D joints, $\mathbf{X}$, and camera parameters, $\mathbf{\Omega}$, encourages smooth motions that are typical of humans in videos.
This is also applied on the 2D keypoint projections, $\mathbf{x}$, as it helps to compensate for spurious errors of the 2D keypoint detector at a particular frame in the video. %

\begin{figure}[t]

\def \imwidth {0.2\linewidth}

\begin{tabular}{ m{0.4cm} cc}
 & 2D joint projection & 3D mesh rendering \\
 No prior & 
 \includegraphics[align=c, width=0.4\linewidth]{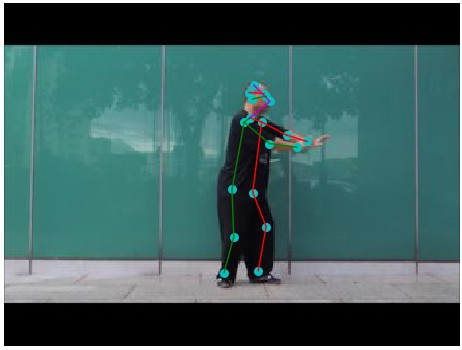} &
 \includegraphics[align=c, width=0.4\linewidth]{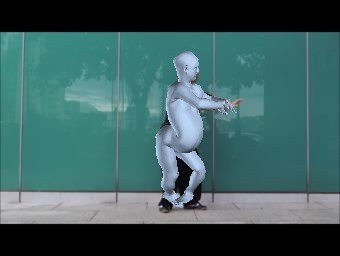}
 \\ \addlinespace[0.1cm]
 With prior &
 \includegraphics[align=c, width=0.4\linewidth]{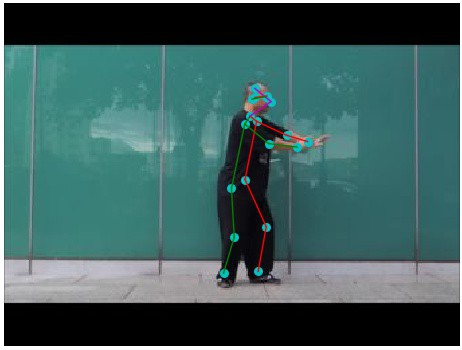} &
 \includegraphics[align=c, width=0.4\linewidth]{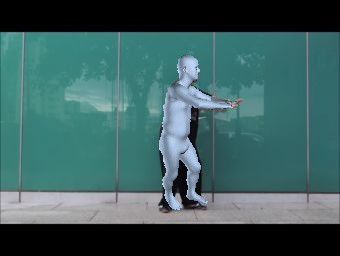}
 \\ \addlinespace[0.1cm]
\end{tabular}
\caption{
Without the prior \eqref{eq:pose_prior}, the SMPL model fit can project well onto 2D keypoints without being in a valid human pose.
}
\label{fig:pose_prior}
\end{figure}

\paragraph{3D Prior:}
There are many 3D poses (including some that are not humanly possible) that project correctly onto the 2D keypoints while also having low temporal error (for example, having all keypoints in a flat plane actually minimises the change with time).  
We use a single $\beta$ for the entire sequence, meaning that changes in distance between 2D keypoints must be explained by pose changes, but telling which keypoint is in front of the other often remains ambiguous.
Therefore, we include a prior term that encourages realistic 3D poses which match the appearance, as illustrated in Fig.~\ref{fig:pose_prior}. 
We use two terms: the same joint angle prior used by \cite{bogo_eccv_2016, huang_3dv_2017, lassner_cvpr_2017}, and another term that robustly encourages the solution to stay close to our initialisation, $(\mathbf{\tilde{\beta}}, \mathbf{\tilde{\theta}})$, which was estimated by the single-frame HMR model.
It is thus defined as:
\begin{align}
E_P(\beta, \theta) &= \sum_t^T E_J(\mathbf{\theta}^t) + \lambda_I E_I(\mathbf{\theta}^t, \mathbf{\beta}) \label{eq:pose_prior} \\
E_J(\mathbf{\theta}) &= 
	-\log \left({\sum_{i}{g_i\mathcal{N}\left(\mathbf{\theta}^{t} ; \mathbf{\mu}_i, \Sigma_i \  \right)}} \right)
\\
E_{I}(\mathbf{\theta}^t, \mathbf{\beta}) &= 
\sum_i^J{ \rho(\mathbf{X}^{t}_i - \mathbf{\tilde{X}}^t_i) } + \lambda_\beta \rho(\mathbf{\beta} - \mathbf{\tilde{\beta}}^{t}).
\label{eq:init_prior}
\end{align}
The joint angle prior, $E_J(\theta)$, is the negative log-likelihood of a Gaussian Mixture Model that was fitted to the joint angles on the CMU Mocap dataset \cite{cmu_mocap}.
$g_i$ are the mixture model weights of 8 Gaussians \cite{bogo_eccv_2016, huang_3dv_2017, lassner_cvpr_2017}, and $\mathbf{\mu}_i$ and $\Sigma_i$ are the mean and covariance of the $i^{th}$ Gaussian.
Multiple modes are used to represent the diverse range of poses which a human can be in.
Note that though our initialisation prior~\eqref{eq:init_prior} penalises deviations in 3D joint positions, these are functions of the SMPL parameters according to~\eqref{eq:smpl}.

\subsection{Optimisation}
\label{sec:optimisation}

We optimise~\eqref{eq:energy} with respect to all SMPL and camera parameters, for all frames in the video, jointly using L-BFGS and Tensorflow.
The solution is first initialised using the results of the per-frame, HMR neural network \cite{kanazawa_cvpr_2018}.
In total there are $10 + 75F$ parameters to be optimised for, where $F$ is the number of frames in the video. 
On a typical clip from Kinetics \cite{kay_arxiv_2017} consisting of 250 frames, the optimisation takes about 8 minutes on a standard CPU or GPU %
(as we did not implement customised kernels for this task), or only 2 seconds per frame.
The time- and memory-efficiency of our method is thus suited for batch, offline processing of videos as done in the following section.

\subsection{Discussion}
As previous works \cite{peng_tog_2018,huang_3dv_2017,zanfir_cvpr_2018,zhang_uist_2018,mehta_tog_2017} have incorporated temporal information into 3D pose estimation using bundle adjustment before, we discuss the differences of our approach:
First, in contrast to \cite{peng_tog_2018,zanfir_cvpr_2018,mehta_tog_2017,zhang_uist_2018}, we use a robust Huber penalty function, and unlike previous approaches, also incorporate additional robustness into our reprojection term for Kinetics data in the next section.
Second, our temporal consistency term is not only on 3D joint positions, but also on 2D joint projections and camera parameters (note that \cite{huang_3dv_2017,zanfir_cvpr_2018,mehta_tog_2017} assume known intrinsics).
Third, unlike previous works, we use our bundle adjustment results to improve a per-frame model.
Fourth, \cite{peng_tog_2018} optimises in the feature space of HMR, whilst we optimise SMPL- and camera-parameters directly.
Additional segmentation masks for model fitting as also used by
\cite{huang_3dv_2017} and \cite{zanfir_cvpr_2018}.

\section{Leveraging Kinetics for weak supervision}
\label{sec:kinetics_training}

Kinetics-400~\cite{kay_arxiv_2017} is a large-scale dataset of human actions collected from YouTube. It contains 400 or more 10s video clips for each of 400 action classes. Each clip is from a different YouTube video, and consequently the dataset contains considerable diversity in people, scenes and camera viewpoints as shown in Fig.~\ref{fig:teaser},\ref{fig:diagram},\ref{fig:pose_prior}. %
We perform bundle adjustment on this dataset to obtain real-world, weakly-supervised training data for 3D pose models.
Bundle adjustment is challenging on Kinetics since there are often multiple people in a frame, shaking cameras, and people are often occluded or move off-camera.
The diversity also results in more frequent failures of our multi-person 2D pose detector \cite{papandreou_cvpr_2017} and HMR \cite{kanazawa_cvpr_2018}.

\paragraph{Dealing with multiple people:}
We could handle multiple people with our formulation in Sec.~\ref{sec:method_single_person} by first tracking a single person through the video, and applying our method to only the tracked region.
However, we found this approach too sensitive to missing detections and tracking failures.
Consequently, we perform tracking to initialise the solution but also augment the per-frame component of our loss function,~\eqref{eq:energy}, to deal with multiple (or potentially no) people, and allow for outliers to be ignored:
\begin{align}
    &E_{R}(\mathbf{\beta}, \mathbf{\theta}^{t}, \mathbf{\Omega}^{t}; \mathbf{x}^{t}_{det, i}) = \\
    &\min\left(\min_{p\in P^t}\sum_{i}^{J}{ w_i h(\mathbf{x}^{t}_{i} - \mathbf{x}^{t,p}_{det, i}}),\tau_{R} \right), \nonumber 
\end{align}
\vspace{-\baselineskip}
\begin{align}
    &E_{I}(\mathbf{\beta}, \mathbf{\theta}^{t}) = \\
    &\min\left(\min_{p\in P^t} \sum_{i}^{J}{ \rho(\mathbf{X}^{t,p}_{i} - \mathbf{\tilde{X}}^t_i) } + \lambda_\beta \rho(\mathbf{\beta} - \mathbf{\tilde{\beta}}^{t}),\tau_{I}\right).   \nonumber
\end{align}
Here, $\tau_{R}$ and $\tau_{I}$ are constants, and $p$ indexes the different person detections $P^t$ in frame $t$.  
Intuitively, the ``inner min'' means that the loss is with respect to the current best-matching 2D pose for each frame.  
However, if estimates from either the 2D pose detection or the HMR model are too far from the current bundle adjustment estimates, they are considered outliers, and the loss is set to a constant (performed  by the ``outer min'').
This means that they no longer affect the bundle adjustment procedure.
There is also substantial jitter in keypoint prediction in Kinetics, due to both 2D detector inaccuracy and camera shake. 
This causes significant problems if a bone is close to parallel with the camera plane: in such cases, jitter in 2D keypoints can often only be explained by large changes in 3D orientation. 
Since we are penalising 3D changes, this encourages the overall algorithm to avoid poses where bones are near parallel with the camera plane.
To mitigate this, we replace the Huber loss, $\rho$, in the reprojection term with a hinge loss, $h$, which is 0 if the error is less than 5 pixels, and behaves like the Huber loss (i.e. L1 error) otherwise.
Finally, to deal with camera motion, we find it advantageous to put an upper bound on the camera translations in~\eqref{eq:temporal}, which is equal to 10\% of the image width, and we do not penalise camera scaling.

\paragraph{Initialisation by tracking:}
The possibility of outliers means that initialisation is important, which we do by first tracking people in 2D using our multi-person pose detector \cite{papandreou_cvpr_2017} that outputs 2D keypoints and bounding boxes for each person in the image.
We select bounding boxes by computing the shortest path 
from the start to the end of the video: distances between detected people in subsequent frames are equal to the mean-squared-error in pixels between detected keypoints.
As there may be missing person detections, we allow the shortest-path algorithm to skip frames with a penalty of $100$ pixels.
Given a selected person detection for each frame, we initialise the 3D pose parameters for each frame using the estimates from  HMR (for any skipped frames, we initialise using the pose from the nearest non-skipped frame).

\paragraph{Training data selection:}
After optimising, we measure the success of the algorithm by the total loss (\ref{eq:energy}).
However, we find that the loss tends to be lowest for people who aren't moving, producing videos that are not suitable to use as training data.
This problem is alleviated by normalising the total loss by the  the 3D trajectory length,
\begin{equation}
    E_{norm}(\mathbf{\beta}, \mathbf{\theta}, \mathbf{\Omega}) = \frac{E(\mathbf{\beta}, \mathbf{\theta}, \mathbf{\Omega}) }{\sum_{t}^{T} \sum_{i}^{J} {\|\mathbf{X}^{t}_{i} - \mathbf{X}^{t-1}_{i}\| }}.
    \label{eq:normalised_energy}
\end{equation}

To obtain training data, we process all videos in Kinetics that do not have more than $6$ detected people in a single frame, as our 2D pose detector and HMR usually fail on crowded scenes.
After running bundle adjustment, we then discard any videos where $E_{norm}$ is above a threshold, retaining roughly 10\% of the original videos.
From these videos, we keep the frames where the 2D reprojections of the 3D poses are inliers with respect to our detected keypoints 
(i.e. $\min_{p\in P^t}\sum_{i}^{J}{ w_i \rho(\mathbf{x}^{t}_{i} - \mathbf{x}^{t,p}_{det, i}})<\tau_R$).

\section{Experiments}

After describing common experimental details in Sec.~\ref{sec:exp_setup}, we first analyse our bundle adjustment method on the Human 3.6M dataset in Sec.~\ref{sec:exp_h36m}.
Although this lab-captured dataset is not particularly realistic, it has metric ground truth 3D which allows us to conduct an ablation study and compare to previous work on 3D pose estimation using the SMPL model.
Thereafter, in Sec.~\ref{sec:exp_kinetics} we run our method large-scale on Kinetics videos before using these predictions in Sec.~\ref{sec:exp_kinetics_training} as weakly-supervised ground truth to retrain a per-frame 3D pose estimation model as described previously in Sec.~\ref{sec:kinetics_training}.

\subsection{Experimental Set-up}
\label{sec:exp_setup}

We initialise the solution to bundle adjustment using the state-of-art HMR neural network \cite{kanazawa_cvpr_2018} which is input an image and outputs SMPL and orthographic camera parameters. 
Unless otherwise specified, we use the publicly released model that has been trained on 3D mocap datasets: Human 3.6M \cite{ionescu_pami_2014} and MPI-3DHP \cite{mehta_3dv_2017}, 2D pose datasets: COCO \cite{lin_eccv_2014}, MPII \cite{andriluka_cvpr_2014} and LSP \cite{johnson_bmvc_2010}, and an adversarial prior that was trained on SMPL model fits using \cite{loper_tog_2014}.
The keypoints that we use for bundle adjustment are obtained using \cite{papandreou_cvpr_2017}, which was trained on the same 2D pose data as HMR and additional data from Flickr collected by the authors.

\subsection{Results on Human 3.6M}
\label{sec:exp_h36m}

Human 3.6M \cite{ionescu_pami_2014} is a popular motion capture dataset and 3D pose benchmark.
Following previous work \cite{pavlakos_cvpr_2017, rogez_cvpr_2017, kanazawa_cvpr_2018}, we downsample the videos from 50fps to 10fps and evaluate on the validation set.
Even so, some videos contain as many as 1175 frames, which we are still able to jointly optimise over.
We report the mean per joint position error (MPJPE) \cite{ionescu_pami_2014},
and also this error after rigid alignment of the prediction with respect to the ground truth using Procrustes Analysis \cite{gower_1975} which we denote as PA-MPJPE.

\begin{table}[tb]
\centering
\caption{Ablation study on Human 3.6M, considering the effect of different terms of our objective function~\eqref{eq:energy}. Mean errors over the validation set are reported.}
\scalebox{0.83}{
\begin{tabular}{lcc}
\toprule
Method		& MPJPE (mm) & PA-MPJPE (mm) \\ \midrule
HMR initialisation \cite{kanazawa_cvpr_2018} & 85.8 & 57.5 \\  
$E_{R}$         		& 154.3 & 99.7   \\ 
$E_{R} + E_{P}$ 		& 79.6 & 55.3   \\ 
$E_{R} + E_{P} + E_{T}$ & 77.8 & 54.3   \\
\midrule 
$E_{R}$ (gt. keypoints)                 & 89.2 & 64.5        \\
$E_{R} + E_{P}$ (gt. keypoints)         & 66.5 & 45.7        \\
$E_{R} + E_{P} + E_{T}$ (gt. keypoints) & 63.3 & 41.6    \\
\bottomrule
\end{tabular}
}
\label{tab:h36m_ablation}
\end{table}

Table \ref{tab:h36m_ablation} shows the effect of the various terms of our objective function in~\eqref{eq:energy}.
We initialise the solution to our bundle adjustment using the public HMR model of \cite{kanazawa_cvpr_2018}, and the error increases if we only use the reprojection error.
As shown in Fig.~\ref{fig:pose_prior}, optimising for reprojection error alone can result in impossible poses.
Note that we are using a single $\beta$ shape parameter across the whole video, but this alone is too weak a constraint.
The addition of the prior term~\eqref{eq:pose_prior} improves results substantially: MPJPE reduces by 6.2mm compared to the HMR initialisation.
Although HMR was also trained with 2D reprojection as one of its loss functions, we obtain better results by explicitly optimising for this term and using HMR 
as an initialisation method.
Note that the 2D pose detector that we use \cite{papandreou_cvpr_2017} has not been trained on Human 3.6M at all.
Our final model, which enforces temporal consistency with not only a single $\beta$ parameter, but smoothness of joints and camera parameters, achieves the best results, significantly improving the MPJPE error of the initial HMR model by 9.4\% and PA-MPJPE by 5.6\%.

The final three rows of Tab.~\ref{tab:h36m_ablation} use ground truth 2D keypoints. %
Note that here, as the ground truth is the projection of 3D joints into the image using the known camera, we have keypoints for occluded joints too.
Each term of our objective function~\eqref{eq:energy} has the same effect on the overall error as before.
However, the MPJPE and PA-MPJPE improve considerably more over the initialisation  of HMR:
Our final model reduces these errors by 26.2 and 27.2\% respectively.
This shows the significant benefits that we can obtain if we have knowledge of occluded keypoints since this further reduces the ambiguity in the problem.

\begin{table}[tb]
\centering
\caption{Comparison of approaches fitting the SMPL model \cite{loper_tog_2015} on Human 3.6M.
We did not use additional Kinetics data here.}
\scalebox{0.82}{
\begin{tabular}{lcc}
\toprule
Method		& MPJPE (mm) & PA-MPJPE (mm)\\ \midrule
Self-Sup \cite{tung_nips_2017} & -- & 98.4 \\
Lassner \etal direct fitting \cite{lassner_cvpr_2017} & -- & 93.9 \\
SMPLify \cite{bogo_eccv_2016} & -- & 82.3 \\
Lassner \etal optimisation \cite{lassner_cvpr_2017} & -- & 80.7 \\
Pavlakos \etal \cite{pavlakos_cvpr_2018} & -- & 75.9 \\
NBF \cite{omran_3dv_2018} & -- & 59.9 \\
MuVS (Note uses 4 cameras) \cite{huang_3dv_2017} & -- & 58.4 \\
HMR \cite{kanazawa_cvpr_2018} & 88.0 & 56.8 \\
\midrule
Ours & \textbf{77.8} & \textbf{54.3} \\
\bottomrule
\end{tabular}
}
\label{tab:h36m_comparison}
\end{table}

\begin{table}[tb]
\centering
\caption{%
Statistics of our bundle-adjustment dataset from Kinetics-$400$.
2D inliers refers to frames where 2D reprojection error was small: $E_R<\tau_R$.
}

\scalebox{1}{
\begin{tabular}{lcc}
\toprule
& Count \\
		\midrule
Total videos & 106 589 \\
Selected videos ($E_{norm}<\tau_R$) & 16 720 \\
\midrule
Total frames in selected videos &  4 141 436 \\
BA inliers &  3 407 686 \\
\bottomrule
\end{tabular}
}
\label{tab:dataset_stats}
\end{table}

Finally, Tab.~\ref{tab:h36m_comparison} shows we achieve the best results on Human 3.6M among other methods utilising the SMPL model.
Note that Mehta \etal \cite{mehta_tog_2017} also perform bundle adjustment to improve the predictions of a CNN model, obtaining an MPJPE of 80.5.
However, as \cite{mehta_tog_2017} do not use the SMPL model, they are not directly comparable.
Additionally, although direct CNN-regression methods such as \cite{dabral_eccv_2018} obtain MPJPE errors of as low as 52.1, they overfit to the Human 3.6M dataset and have been shown to be significantly outperformed by SMPL-based approaches on real-world datasets such as 3DPW \cite{von_marcard_eccv_2018} by Kanazawa \etal \cite{kanazawa_cvpr_2019}.

\subsection{Results on Kinetics}
\label{sec:exp_kinetics}
Given that our algorithm can reliably improve 3D estimates, we apply our method to a large-scale video dataset to produce training data for single-frame 3D pose estimation.
We used the entirety of Kinetics-400 \cite{kay_arxiv_2017} (400+ clips of 400 action classes), after automatically selecting videos as described in Sec.~\ref{sec:kinetics_training}.

Table \ref{tab:dataset_stats} shows the statistics of the important stages in this process. 
We first pre-select about 16.7K videos based on the normalized bundle adjustment loss \eqref{eq:normalised_energy}, resulting in 4.1M frames. 
The bundle adjustment matched the prediction of the 2D pose detector \cite{papandreou_cvpr_2017} for 3.4M out of 4.1M frames (we used a threshold of $\tau_R=50$ pixels total error to determine outliers). 
Visual inspection showed that the 3D pose detector was fairly reliable: for the majority of outlier frames, the person was occluded or had simply left the frame.

\begin{table}[tb]
\caption{The most common action classes of the videos selected from Kinetics.
Our bundle adjustment method works well on action classes that do not appear in motion capture datasets, e.g., those that occur outdoors or contain multiple people. %
}
\centering
\scalebox{1}{
\begin{tabular}{lcc}
\toprule
Action class & Selected videos & Selected frames \\ \midrule
Roller skating      	& 259 &  55 941 \\
Hula hooping 			& 247 &  56 498 \\
Salsa dancing       	& 229 &  50 377 \\
Spinning poi   			& 200 &  42 316 \\
Dancing ballet          & 199 &  44 016 \\
Playing drums       	& 193 &  41 318 \\
Tap dancing 			& 192 &  44 757 \\

\bottomrule
\end{tabular}
}
\label{tab:dataset_classes}
\vspace{-\baselineskip}
\end{table}

Table \ref{tab:dataset_classes} lists the action classes from Kinetics that were selected most often, showing that none of them appear in existing mocap datasets \cite{ionescu_pami_2014, sigal_ijcv_2010, mehta_3dv_2017}.
Mocap datasets only contain actions performed by a single person, in contrast to classes such as ``tap dancing'' and ``salsa dancing'' which bundle adjustment performs well on.
Similarly, our method is effective on outdoor activities such as ``roller skating'' and ``spinning poi'' which cannot be recorded by mocap.
There were no classes without any selected videos, but for several classes (e.g., ``knitting'' and ``tying tie''), where a person is rarely fully visible, we only selected 1 video each.
Some qualitative examples of the diversity of our dataset are shown in Fig.~\ref{fig:kinetics_diversity}.
All experimental hyperparameters are included in the appendix.

\begin{figure*}

\def \imheight {2.5cm}
\setlength{\tabcolsep}{1pt} %
\hspace*{-0.25cm}
\begin{tabular}{cccccccccccc}
\includegraphics[height=\imheight]{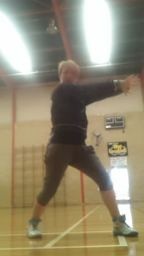} & 
\includegraphics[height=\imheight]{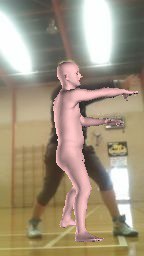} &
\includegraphics[height=\imheight]{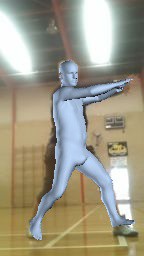} &
\includegraphics[height=\imheight]{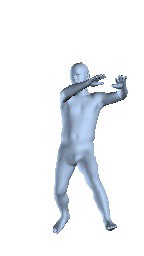} &
\includegraphics[height=\imheight]{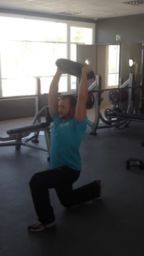} & 
\includegraphics[height=\imheight, trim={0cm 0 0cm 0},clip]{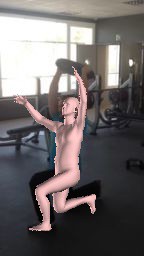} &
\includegraphics[height=\imheight,trim={0cm 0 0cm 0},clip]{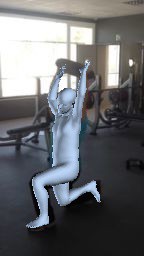} &
\includegraphics[height=\imheight,trim={0cm 0 0cm 0},clip]{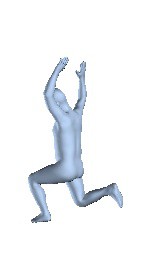} &
\includegraphics[height=\imheight]{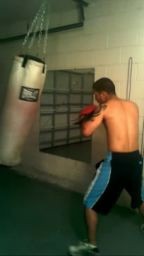} & 
\includegraphics[height=\imheight]{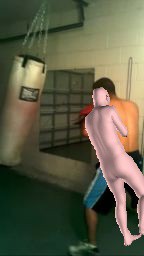} &
\includegraphics[height=\imheight]{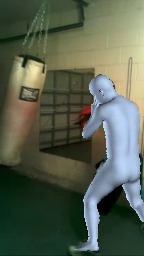} &
\includegraphics[height=\imheight]{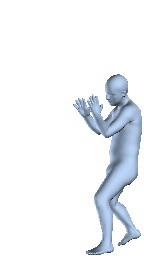}
\\
\end{tabular}

\def \imheight {2.1cm}
\hspace*{-0.30cm}
\setlength{\tabcolsep}{1pt} %
\begin{tabular}{cccccccc}
\includegraphics[height=\imheight]{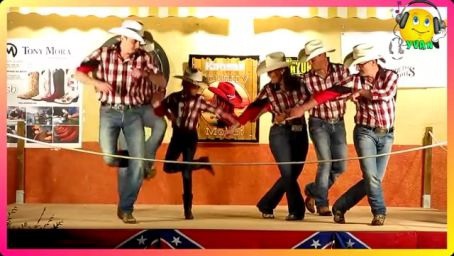} & 
\includegraphics[height=\imheight, trim={1.5cm 0 7.5cm 0},clip]{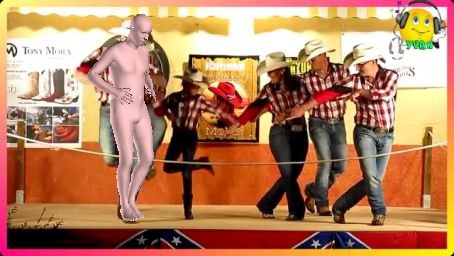} &
\includegraphics[height=\imheight, trim={1.5cm 0 7.5cm 0},clip]{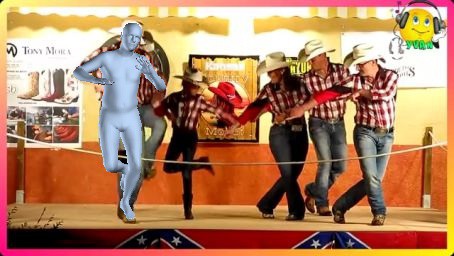} &
\includegraphics[height=\imheight, trim={1.5cm 0 8.5cm 0},clip]{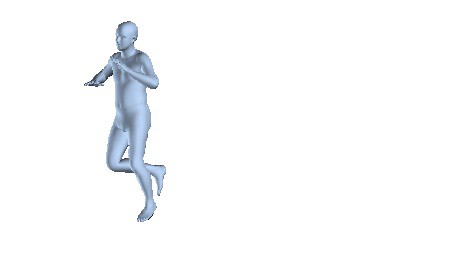} &
\includegraphics[height=\imheight, trim={1.9cm 0 1.9cm 0},clip]{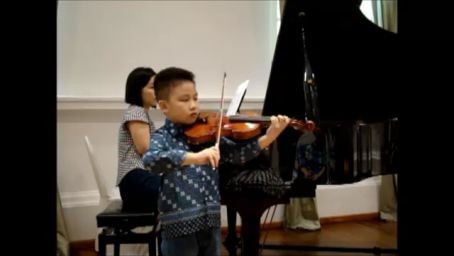} & 
\includegraphics[height=\imheight, trim={2cm 0 5cm 0},clip]{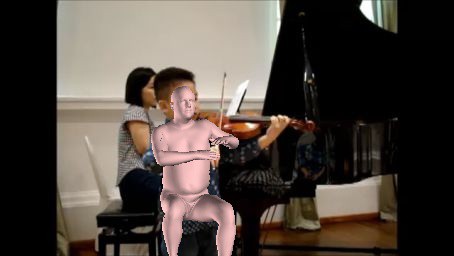} & 
\includegraphics[height=\imheight, trim={2cm 0 5cm 0},clip]{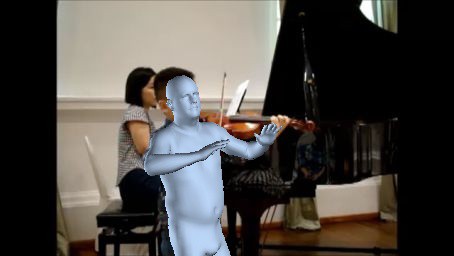} &
\includegraphics[height=\imheight, trim={2cm 0 5cm 0},clip]{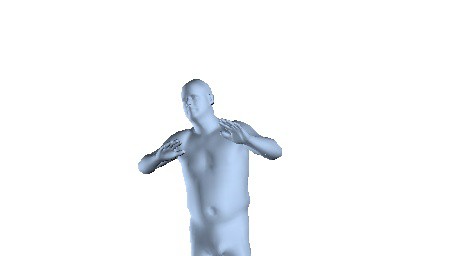} 
\\
\end{tabular}

\def \imheight {2.1cm}
\hspace*{-0.25cm}
\begin{tabular}{cccccccc}
\includegraphics[height=\imheight]{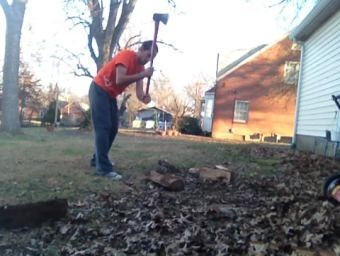} & 
\includegraphics[height=\imheight, trim={1cm 0 2cm 0},clip]{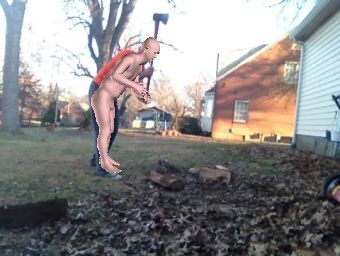} &
\includegraphics[height=\imheight, trim={1cm 0 2cm 0},clip]{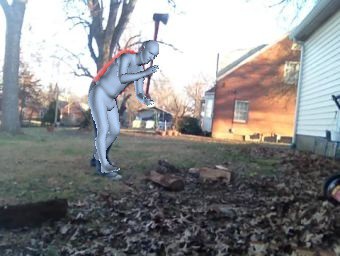} &
\includegraphics[height=\imheight, trim={1.5cm 0 4.4cm 0},clip]{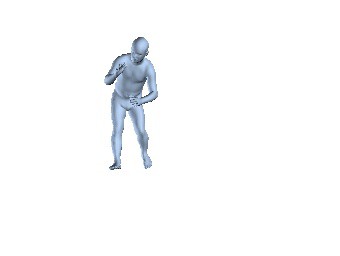} &
\includegraphics[height=\imheight, trim={1.9cm 0 1.9cm 0},clip]{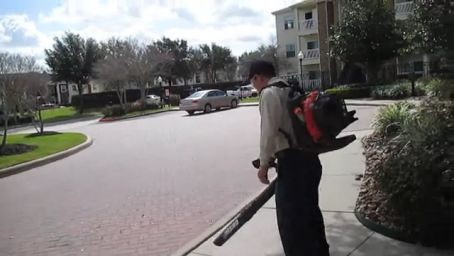} & 
\includegraphics[height=\imheight, trim={5cm 0 2cm 0},clip]{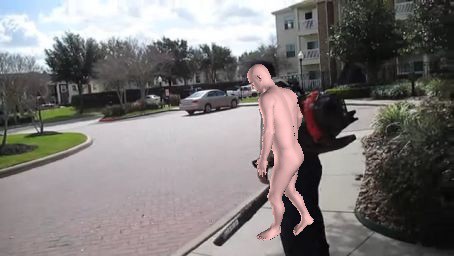} & 
\includegraphics[height=\imheight, trim={5cm 0 2cm 0},clip]{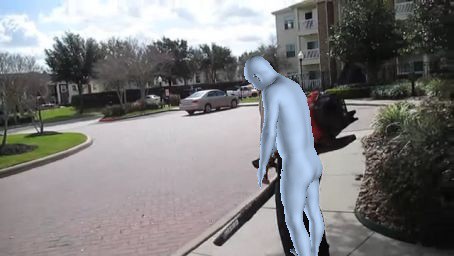} &
\includegraphics[height=\imheight, trim={5cm 0 2cm 0},clip]{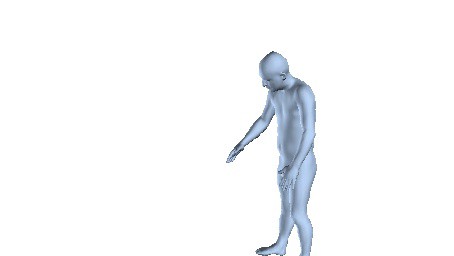}
\\
\end{tabular}

\def \imheight {2.1cm}
\hspace*{-0.25cm}
\begin{tabular}{cccccccc}
\includegraphics[height=\imheight]{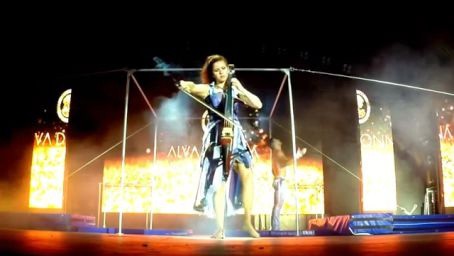} & 
\includegraphics[height=\imheight, trim={4.5cm 0 4.5cm 0},clip]{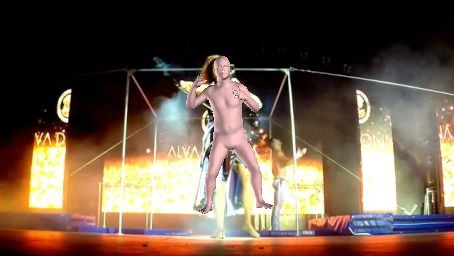} & 
\includegraphics[height=\imheight, trim={4.5cm 0 4.5cm 0},clip]{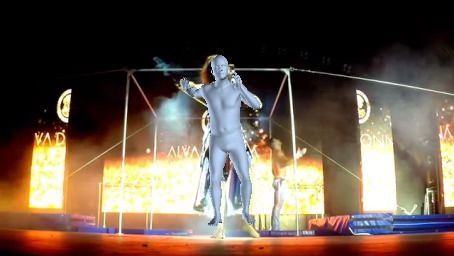} & 
\includegraphics[height=\imheight, trim={5cm 0 5cm 0},clip]{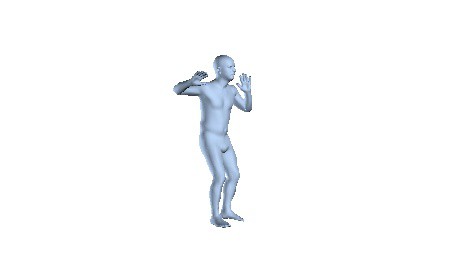} & 
\includegraphics[height=\imheight]{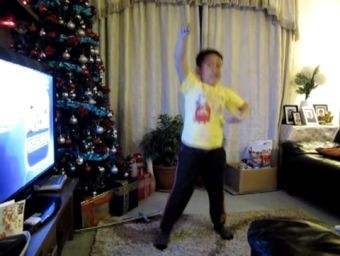} &
\includegraphics[height=\imheight, trim={1.6cm 0 1.1cm 0},clip]{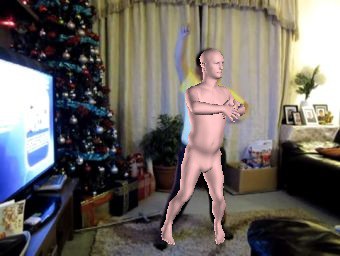} & 
\includegraphics[height=\imheight, trim={1.6cm 0 1.1cm 0},clip]{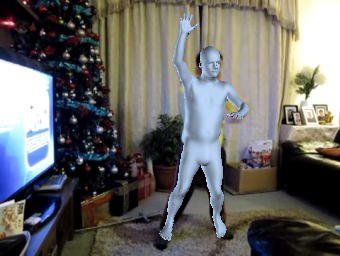} &
\includegraphics[height=\imheight, trim={1.6cm 0 1.1cm 0},clip]{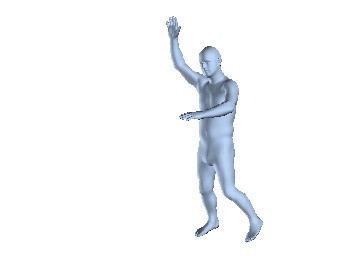}
\\
\end{tabular}

\caption{The dataset we automatically generated from Kinetics has a diverse range of scenes, people, camera viewpoints and action classes not found in motion capture.
We show the input frame, results for a single tracked person (which are cropped for display) of HMR (pink) and bundle adjustment (blue), and the bundle adjustment result from another view respectively.
Note how bundle adjustment typically improves the per-frame estimates of HMR.
}
\label{fig:kinetics_diversity}
\end{figure*}

\setlength{\tabcolsep}{6pt} %

\subsection{Weak supervision from Kinetics}
\label{sec:exp_kinetics_training}

We utilise the training data that we automatically obtained in the previous section to retrain a new HMR model from Imagenet initialisation.
We use the original training data (described in Sec.~\ref{sec:exp_setup}) too, and use a model only trained on this data as our baseline.
We evaluate on the recently released 3D Poses in the Wild dataset (3DPW) \cite{von_marcard_eccv_2018} in Sec.~\ref{sec:exp_3dpw}, which consists of outdoor videos captured in real-world conditions, HumanEVA \cite{sigal_ijcv_2010}, a mocap dataset in Sec.~\ref{sec:exp_humaneva} and Ordinal Depth \cite{pavlakos_cvpr_2018b} which provides a good proxy task for 3D pose esitmation on unconstrained, real-world internet images.
Our network has never been trained on images from either of these datasets.
To verify the effect of Kinetics training, we trained a model with all frames from our automatically-generated dataset (Kinetics 3M), and also with a random subset of 10\% of the frames in our dataset (Kinetics 300K).

When retraining the HMR model on Kinetics data, we made modifications to the HMR training procedure \cite{kanazawa_cvpr_2018}.
These are detailed in Sec.~\ref{sec:exp_kinetics_ablation}, where we also show that our modifications only help for training on Kinetics data, and not when using only the original training data used by HMR.

\begin{table}[t]
	\caption{Results on the 3DPW \cite{von_marcard_eccv_2018}, HumanEVA \cite{sigal_ijcv_2010} and Ordinal Depth \cite{pavlakos_cvpr_2018b} datasets when training with our Kinetics datasets.
	We evaluate the HMR model retrained by us on its original training data using the author's public code, and the HMR model trained on its original data and 300K and 3M frames from our Kinetics dataset.
	For 3DPW and HumanEVA, we report the PA-MPJPE error in mm (lower is better), and for Ordinal Depth, we report the accuracy in \% (higher is better).
	}
	\scalebox{0.76}{
		\begin{tabularx}{1.3\linewidth}{lYYY}
			\toprule
			Dataset                        & Original data & Original + Kinetics 300K & Original + Kinetics 3M \\ \midrule
			3DPW ($\downarrow$)            & 77.2 & 73.8 & \textbf{72.2} \\
			HumanEVA ($\downarrow$)          & 85.7 & 83.5 & \textbf{82.1} \\
			\midrule 
			Ordinal depth ($\uparrow$)    &  82.5 & 83.7 & \textbf{84.6} \\
			\bottomrule
		\end{tabularx}
	}
	\label{tab:kinetics_train_results}
\end{table}

\subsubsection{3D Poses in the Wild}
\label{sec:exp_3dpw}

This recently released dataset contains 60 clips, consisting of outdoor videos captured from a moving mobile phone and 17 IMUs attached to the subjects \cite{von_marcard_eccv_2018}.
The IMU data allowed the authors to accurately compute 3D poses which we use as ground truth.
We evaluate on the test set comprising 24 videos, using the 14 keypoints that are common across both MS-COCO and SMPL skeletons, as also done by \cite{kanazawa_cvpr_2019}.
We only evaluate on frames where enough of the person is visible to estimate a 3D pose for it.
This is performed by discarding examples where less than 7 ground-truth 2D keypoints are visible.
We compute the Procrustes-aligned error independently for each pose, and then average errors for each tracked person within each video, before finally averaging over the entire dataset (thus videos with two people count twice as much as videos with one).

Table~\ref{tab:kinetics_train_results} shows how using additional data from Kinetics improves results on this dataset.
Training with 300K frames of Kinetics data improves the PA-MPJPE by 3.4mm, and our model trained with all 3M frames of Kinetics improves further by 5 mm over the baseline.
Our Kinetics-trained model also outperforms the public HMR model \cite{kanazawa_cvpr_2018} (trained by the authors) which obtains a PA-MPJPE error of 74.9.
While isolated checkpoints from our reimplementation of HMR perform as well as the public model, not all do; Tab~\ref{tab:kinetics_train_results} computes the mean of 20 checkpoints (roughly 1500 training iterations apart) to minimise variance.

\subsubsection{HumanEVA}
\label{sec:exp_humaneva}

HumanEVA \cite{sigal_ijcv_2010} is an indoor motion-capture dataset where we follow the evaluation protocol of \cite{bogo_eccv_2016} on the validation set.
Although HumanEVA does not contain ``in the wild'' data, it is a dataset which our HMR model has not been trained on at all.
Table~\ref{tab:kinetics_train_results} shows how adding additional data from our Kinetics dataset improves performance on this dataset compared to our baselines that were trained without Kinetics.
Our model trained with 300K frames of Kinetics data improves the PA-MPJPE by 2.2 mm, and the model trained with 3M Kinetics frames improves further by 3.6 mm over our baseline.
The public HMR model obtains a PA-MPJPE error of 83.5, which is also worse than our Kinetics-trained model.

\subsubsection{Ordinal Depth}

A key challenge with 3D pose estimation in-the-wild is the lack of ground truth for people performing arbitrary, unconstrained actions in-the-wild (as typically found on images scraped from the internet).  
However, a suitable proxy for 3D pose estimation quality is ordinal depth \cite{taylor_cviu_2000, pavlakos_cvpr_2018b} -- \ie given two keypoints, predict the relative depth ordering by specifying which keypoint is in front of the other.
This utility of this task was demonstrated by Taylor \cite{taylor_cviu_2000}, who showed that the 3D skeleton of a person could be reconstructed perfectly if exact 2D keypoint correspondences, bone lengths and ordinal relations between keypoints were known, assuming an orthographic camera.

Although humans cannot annotate 3D pose or absolute depth, they can reliably label ordinal depth \cite{pavlakos_cvpr_2018b}.
We thus evaluate on the Ordinal Depth dataset \cite{pavlakos_cvpr_2018b} which added ordinal depth annotations to the MPII \cite{andriluka_cvpr_2014} and LSP \cite{johnson_bmvc_2010} 2D pose datasets of real-world internet images.
We evaluate on 2606 images from the validation sets of MPII and LSP, as images from the training set were used to train HMR (we do not use any of the ordinal depth information during training).
For each person, each pair of keypoints is labelled either ``in front'', ``behind'', or ``ambiguous''.
To evaluate, we compute the 3D pose for each person and then obtain ordinal depth for each pair of keypoints.
We report the average accuracy, ignoring keypoint pairs labelled as ambiguous.

Table~\ref{tab:kinetics_train_results} shows the benefits we get for this task by training on Kinetics.
Even a relatively small amount of Kinetics data provides noticeable improvements on this dataset, with further benefits from our entire dataset.
As expected, training on real-world data from Kinetics helps on ordinal depth predictions of real-world images.

\paragraph{} These experiments thus show how we can effectively
use Kinetics data to improve the per-frame HMR model on
multiple datasets.
We also achieve greater improvements on the real-world 3DPW dataset, compared to the mocap HumanEVA dataset.

\subsubsection{HMR training modification and ablation}
\label{sec:exp_kinetics_ablation}

\begin{table}[t]
\caption{Ablation study of our HMR retraining schemes. 
	PA-only 3D means during our retraining of HMR, we discard the losses on SMPL joints and absolute 3D locations and only use losses on joints after Procrustes alignment.  
	No 2D means disabling all HMR datasets that contain only 2D data (and therefore disabling the adversarial prior which is only used on 2D datasets).
	}	
\scalebox{0.76}{
\begin{tabularx}{1.3\linewidth}{Xcc}
\toprule
  & 3DPW & HumanEVA 
\\ \midrule
Original data, original training 			&  77.2  & 85.7  \\
Original data, PA-only 3D        			&  78.7  & 86.2  \\ 
Original data, PA-only 3D, no 2D 			&  144.6 & 99.2  \\
\midrule
Original + Kinetics data, original training	&  91.1& 90.0 \\
Original + Kinetics data, PA-only 3D, no 2D & \textbf{72.2}& \textbf{82.1} \\
\bottomrule
\end{tabularx}
\label{tab:kinetics_train_ablation}
}
\end{table}

When training with Kinetics data, we find that it is beneficial to not use any of the original 2D data used by HMR, and thus also to not use the adversarial pose prior since it is only used on 2D pose datasets \cite{kanazawa_cvpr_2018}.
We suspect that this is because the adversarial pose prior encourages predictions that are closer to the mean pose, and since we use HMR to initialise our bundle adjustment, our Kinetics data may also have a slight bias towards this mean pose.
Applying the same prior while retraining may aggravate this problem.

We also find it's important to train only on 3D keypoints after Procrustes alignment, rather than training directly on SMPL joint angles and absolute 3D keypoint locations.
Note this means that HMR only learns to predict the camera orientation by minimizing 2D reprojection error.  
We suspect that this strategy is effective because Kinetics has a very large range of camera orientations, which may not match well with evaluation datasets that have less variety in camera pose.

Table~\ref{tab:kinetics_train_ablation} shows that our modifications to the HMR training procedure help only when we train with additional Kinetics data.
When using the original training data, our modified training procedure does not improve results.
Removing the original 2D data from training also has a large negative impact on performance.
This is because the original training data has a relatively small amount of 3D supervision (Human 3.6M \cite{ionescu_pami_2014} and MPI-3DHP \cite{mehta_3dv_2017}).

\section{Conclusion and Future Work}
We presented a bundle-adjustment algorithm to leverage the temporal context in a video in order to improve estimates of the 3D pose of a person.
Furthermore, we applied this to YouTube videos from Kinetics and automatically generated a dataset which we used to improve per-frame 3D pose estimators, demonstrating how we can effectively use large amounts of unlabelled data to improve existing models.

Bundle adjustment was effective because videos are shot in a 3D world where people move slowly (relative to the camera framerate), and the person's size and appearance remain consistent over time.  
If properly characterised, these constraints can give strong supervision to algorithms, which allows us to break out of the environments which motion capture devices are restricted to.
We believe there is far more 3D structure to exploit, because people don't behave in a vacuum.
People act under gravity, are supported by ground planes and interact with objects.
Therefore, we aim in future to use physical constraints and information about human actions to constrain poses and predict the objects that people are interacting with to estimate their affordances.

\paragraph{Acknowledgements:}
We would like to thank Jean-Baptiste Alayrac, Relja Arandjelovi\'c, Jo\~ao Carreira, Rohit Girdhar, Viorica P{\u a}tr{\u a}ucean and Jacob Walker for valuable discussions.

{\small
\bibliographystyle{ieee}
\bibliography{bibliography}
}

\clearpage
\appendix

\section*{Appendix}
Section \ref{sec:exp_details} lists the hyperparameters we used for our bundle adjustment, whilst Sec.~\ref{sec:dataset_stats} provides some more details about the dataset we automatically generated from Kinetics.
\section{Experimental Details}
\label{sec:exp_details}

Table~\ref{tab:hyperparam_values} shows the values of our bundle adjustment hyperparameters for our experiments.

\begin{table}[h]
	\centering
	\caption{Bundle adjustment hyperparameters used for experiments
	}
	\begin{tabular}{ccc}
	\toprule
	Hyperparameter & Human 3.6M \cite{ionescu_pami_2014} & Kinetics \cite{kay_arxiv_2017} \\ 
	\midrule
	$\lambda_R$         & $1 \times 10^{-3}$    & $1 \times 10^{-3}$ \\
	$\lambda_I$         & $10$    				& $0.2$ \\ %
	$\lambda_{\beta}$   & $0.2$    				& $0.05$ \\
	$\lambda_{J}$       & $1  \times 10^{-4}$   & 1  $\times 10^{-4}$ \\
	$\lambda_1$         & $5$    				& $0.2$ \\ 
	$\lambda_2$         & $1 \times 10^{-4}$    & $1 \times 10^{-3}$ \\ %
	$\lambda_3$         & $2$    & $20$\\
	$\tau_R$            & --    & $50$\\
	$\tau_I$            & --    & $2 \times 10^{-2}$ \\
  \bottomrule
	\end{tabular}
\label{tab:hyperparam_values}
\end{table}

Note that the 2D joint positions, $\mathbf{x}$ are measured in pixels, and that the largest spatial dimensions of a video frame is typically around $450$.
On the other hand, the 3D joint positions, $\mathbf{X}$ and camera parameters are typically in the range $[-1, 1]$.
As the range of the 2D joint positions is higher, the values of $\lambda_R$ and $\lambda_2$, are small, even though they have a significant effect on the bundle adjustment.

$\lambda_I$ and $\lambda_{\beta}$ are higher on Human 3.6M than they are on Kinetics.
These weights are used in the prior term that encourages the bundle adjustment result to stay close to the initialisation (Eq.~8 of main paper). 
The initialisation that we get from HMR \cite{kanazawa_cvpr_2018} is far better on Human 3.6M than on Kinetics, which is why $\lambda_I$ and $\lambda_{\beta}$ are higher on Human 3.6M.
It is expected that HMR performs better on Human 3.6M as it has been trained with 3D supervision from this dataset.
\section{Dataset statistics}
\label{sec:dataset_stats}

\begin{figure*}
	\centering
	\vspace{-2\baselineskip}
	\begin{tabularx}{\linewidth}{Y}
	 	\includegraphics[width=0.5\linewidth]{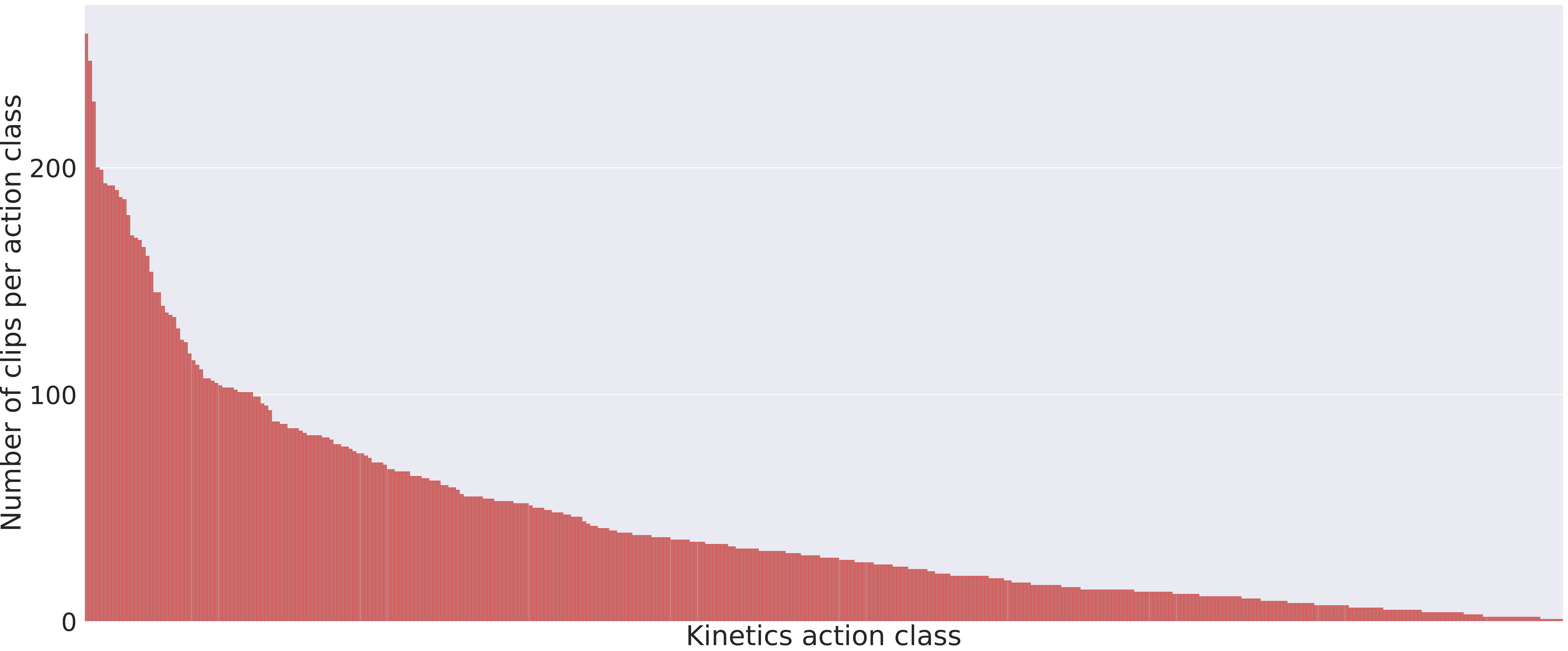} \\ 
	 	(a) Number of clips selected per action class. For legibility, the action classes are not shown in the x axis, and the most- and least-common classes are shown below instead. \\
	 	\includegraphics[width=0.5\linewidth]{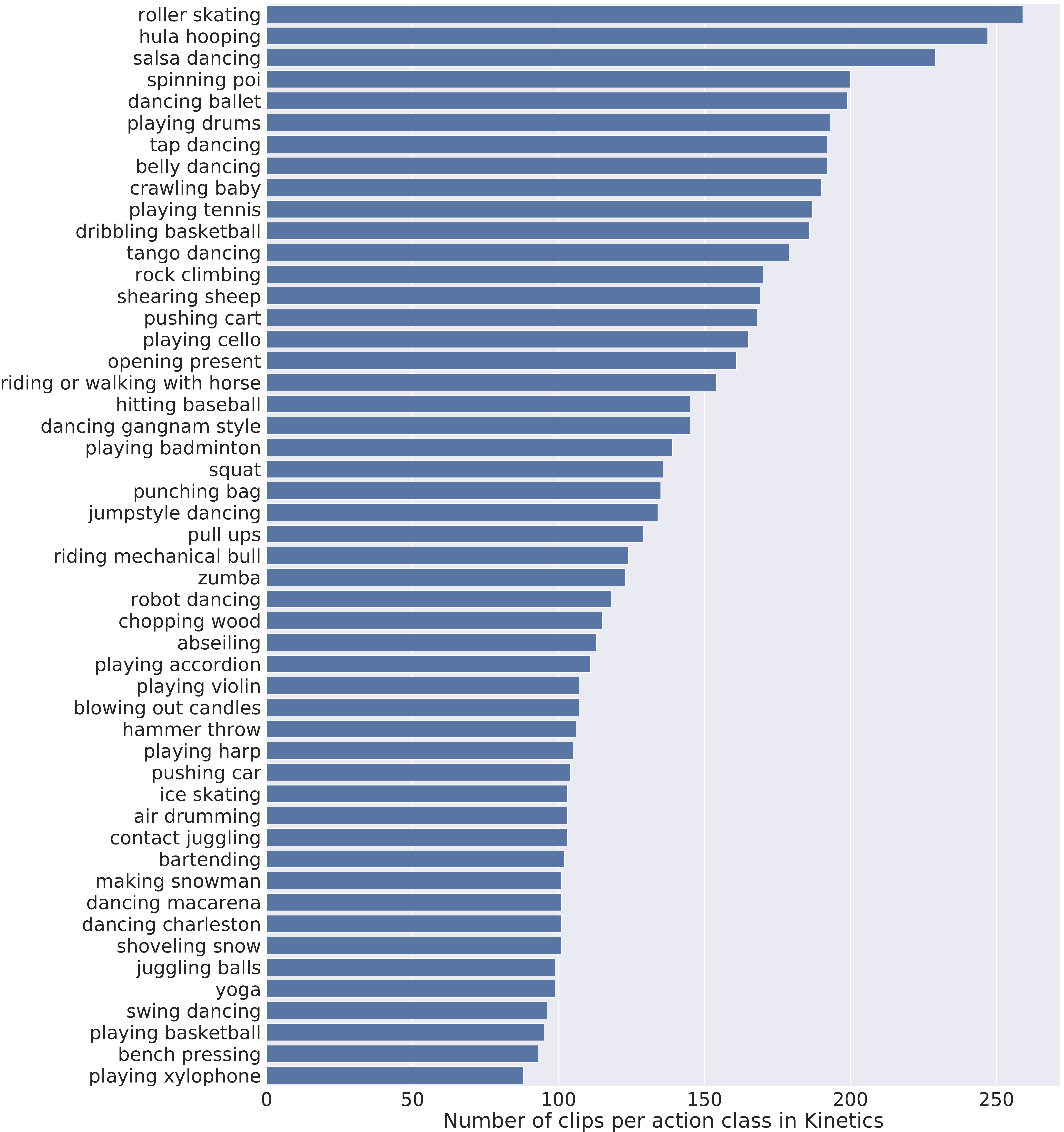} \\ 
	 	(b) The number of clips selected per class for the 50 most common Kinetics action classes. \\
	 	\includegraphics[width=0.5\linewidth]{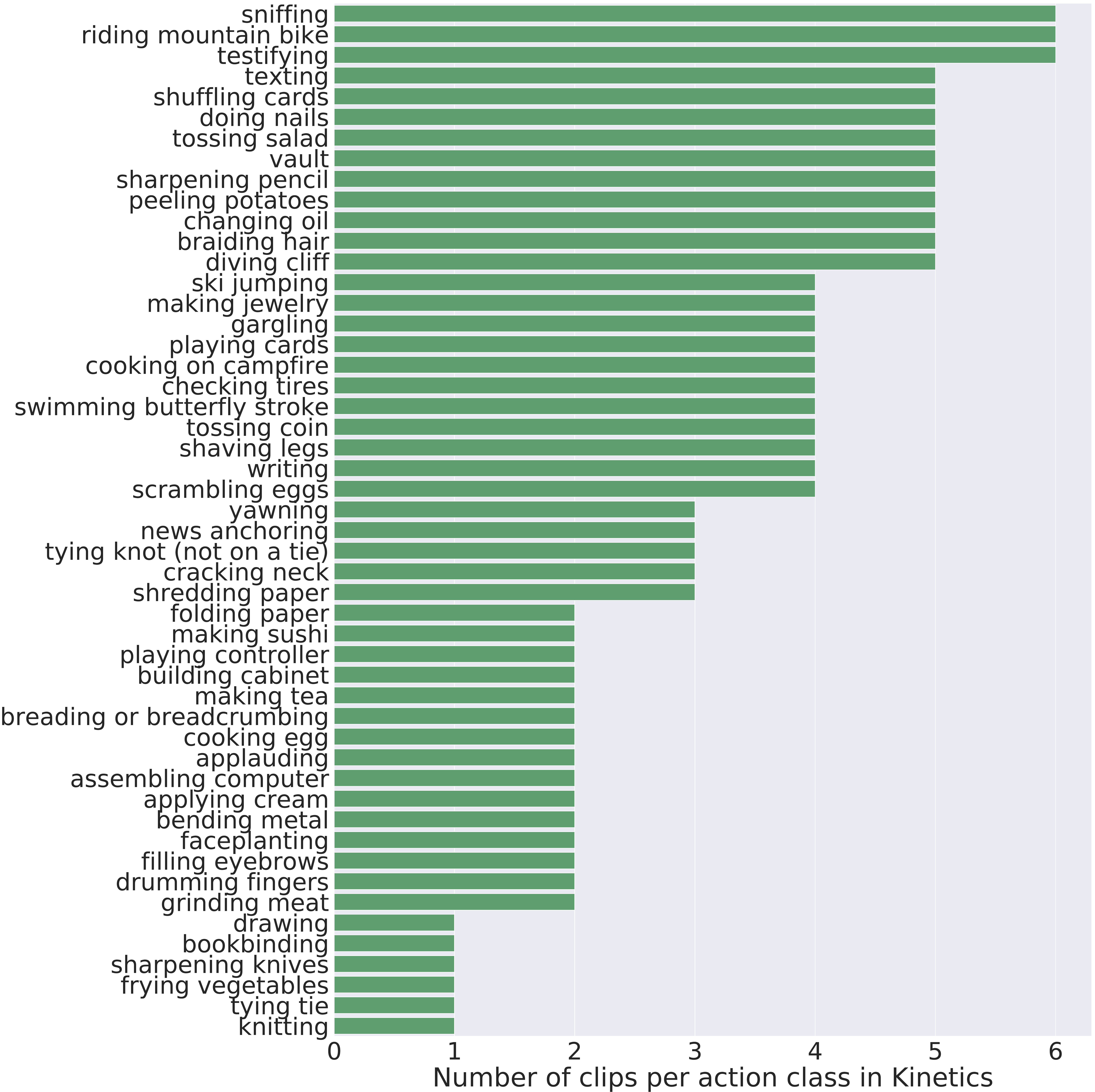} \\ 
	 	(c) The number of clips selected per class for the 50 least common Kinetics action classes. \\
	\end{tabularx}
	\vspace{\baselineskip}
	\caption{Number of video clips selected per action class in the Kinetics dataset. (a) shows the overall distribution of video clips selected per action class, whilst (b) and (c) show the most- and least-common Kinetics action classes respectively.}
	\label{fig:dataset_stats}

\end{figure*}

Figure~\ref{fig:dataset_stats} visualises the distribution of Kinetics action classes in our dataset.
We can see that the distribution has a fairly long tail:
Our bundle adjustment method works well for a variety of object classes, including many types of dancing and various outdoor activities, where there are usually not many people in the video clip and the whole body is visible.
There are also many classes for which only a handful of videos are automatically selected. These are typically classes such as ``tying tie'', ``bookbinding'' and ``knitting'' where the person is usually not fully visible.
Note that there are 400+ clips for each action in the Kinetics-400 dataset \cite{kay_arxiv_2017} that we use, and that we have always selected at least one video of each action class.

\end{document}